\documentclass[conference]{IEEEtran}
\IEEEoverridecommandlockouts

\usepackage{cite}
\usepackage{amsmath,amssymb,amsfonts}
\usepackage{algorithmic}
\usepackage{graphicx}
\usepackage{textcomp}
\usepackage{xcolor}
\def\BibTeX{{\rm B\kern-.05em{\sc i\kern-.025em b}\kern-.08em
    T\kern-.1667em\lower.7ex\hbox{E}\kern-.125emX}}

\usepackage[utf8]{inputenc}
\usepackage{algorithm}
\usepackage{epsfig}
\usepackage{mathtools}
\usepackage{multirow}
\usepackage{subcaption}
\usepackage{lipsum}
\usepackage{tablefootnote}
\usepackage{url}
\newcommand{\ie}{\textit{i.e.}}
\newcommand{\eg}{\textit{e.g.}}

\makeatletter
\newcommand{\linebreakand}{%
  \end{@IEEEauthorhalign}
  \hfill\mbox{}\par
  \mbox{}\hfill\begin{@IEEEauthorhalign}
}
\makeatother

\newcommand\blfootnote[1]{%
  \begingroup
  \renewcommand\thefootnote{}%
  \addtocounter{footnote}{-1}%
  \footnote{\noindent\hspace{-1em}\rule{0.2\textwidth}{0.4pt}\par #1}%
  \endgroup
}

\begin{document}

\title{Sample Efficient Reinforcement Learning via \\ Large Vision Language Model Distillation\\

}

\author{\IEEEauthorblockN{Donghoon Lee$^\dagger$}
\IEEEauthorblockA{\textit{Robotics Program} \\
\textit{KAIST}\\
Daejeon, South Korea \\
dh\_lee99@kaist.ac.kr}
\and

\IEEEauthorblockN{Tung M. Luu$^\dagger$}
\IEEEauthorblockA{\textit{Electrical Engineering} \\
\textit{KAIST}\\
Daejeon, South Korea \\
tungluu2203@kaist.ac.kr}
\and

\IEEEauthorblockN{Younghwan Lee}
\IEEEauthorblockA{\textit{Electrical Engineering} \\
\textit{KAIST}\\
Daejeon, South Korea \\
youngh2@kaist.ac.kr}
\and

\IEEEauthorblockN{Chang D. Yoo}
\IEEEauthorblockA{\textit{Electrical Engineering} \\
\textit{KAIST}\\
Daejeon, South Korea \\
cd\_yoo@kaist.ac.kr}
}

\maketitle

\blfootnote{$\dagger$ The authors are equally contributed.

This work was supported by Institute for Information \& communications Technology Planning \& Evaluation (IITP) grant funded by the Korea government(MSIT) (No.RS-2021-II211381, Development of Causal AI through Video Understanding and Reinforcement Learning, and Its Applications to Real Environments).} %
\begin{abstract}

Recent research highlights the potential of multimodal foundation models in tackling complex decision-making challenges. However, their large parameters make real-world deployment resource-intensive and often impractical for constrained systems. Reinforcement learning (RL) shows promise for task-specific agents but suffers from high sample complexity, limiting practical applications. To address these challenges, we introduce LVLM to Policy (LVLM2P), a novel framework that distills knowledge from large vision-language models (LVLM) into more efficient RL agents. Our approach leverages the LVLM as a teacher, providing instructional actions based on trajectories collected by the RL agent, which helps reduce less meaningful exploration in the early stages of learning, thereby significantly accelerating the agent's learning progress. Additionally, by leveraging the LVLM to suggest actions directly from visual observations, we eliminate the need for manual textual descriptors of the environment, enhancing applicability across diverse tasks. Experiments show that LVLM2P significantly enhances the sample efficiency of baseline RL algorithms.  The code is available at \texttt{https://github.com/i22024/LVLM2P}

\end{abstract}

\vspace{-0.7mm} 
\section{Introduction}
\vspace{0.3mm}
Reinforcement learning (RL) has rapidly become a leading approach for tackling complex decision-making problems, achieving success in domains such as gaming \cite{silver2016mastering}, autonomous driving \cite{you2019advanced}, and robotics \cite{kalashnikov2018qt}. However, despite these achievements, RL algorithms face a critical limitation in sample inefficiency, particularly in high-dimensional and sparse reward environments, requiring extensive interactions with the environment to train agents effectively \cite{schrittwieser2020mastering,luu2021hindsight,Luu2022SampleEfficiency,luu2024predictive,berner2019dota}. This need for extensive data imposes substantial demands on human supervision, safety checks, and resets \cite{garcia2015comprehensive,atkeson2015no}, which can hinder the deployment of RL algorithms in real-world environments where such comprehensive training is often unfeasible \cite{arulkumaran2017deep}. 

At the same time, multimodal foundation models (FMs) trained on massive, Internet-scale datasets have demonstrated exceptional capabilities as general-purpose agents, excelling in various complex reasoning tasks \cite{radford2019language,brown2020language,bommasani2021opportunities,reid2024gemini,openai2023gpt4v}. Recent works have attempted to leverage large vision-language models (LVLMs) as planners \cite{ahn2022can,huang2022language}, or fine-tune LVLMs for direct use as policies \cite{szot2023large,zhai2024fine}. These approaches enable agents to tap into vast general knowledge about the world, allowing them to learn and solve tasks more efficiently. However, despite their potential, FMs typically contain hundreds of billions of parameters \cite{touvron2023llama,chowdhery2023palm}, making them highly impractical for real-time systems, especially in resource-constrained devices. The large size of these models requires substantial computational resources, including significant memory and processing power, which are often unavailable in such systems. Moreover, relying on commercial FMs \cite{reid2024gemini,openai2023gpt4v} to offload computational workloads introduces additional cost challenges, as these services typically charge based on usage, further limiting their feasibility for long-term deployment on constrained systems. The question then arises: can we design a compact agent that quickly acquires the capabilities of multimodal FMs while maintaining efficiency and adaptability for specific tasks in resource-constrained systems?

In addressing this challenge, we introduce \textbf{LVLM} to \textbf{P}olicy (LVLM2P), a novel framework that distills knowledge \cite{hinton2015distilling} from a pretrained LVLM into a compact student agent tailored for a specific task. Specifically, during training, the student agent collects a trajectory and queries the LVLM, prompted with few-shot examples, for instructional actions based on visual observations within the trajectory. The agent is then trained to imitate the LVLM's behavior by optimizing a distillation objective alongside the conventional RL objective. This training paradigm enables the student agent to not only leverage guidance from the LVLM but also learn from its own online interactions with the environment, allowing it to identify and correct any mistakes made by its teacher, ultimately leading to improved performance on the target task. During testing, the student agent operates independently without further interaction with the LVLM. 

\begin{figure*}[ht]
    \centering
    \includegraphics[width=0.89\textwidth]{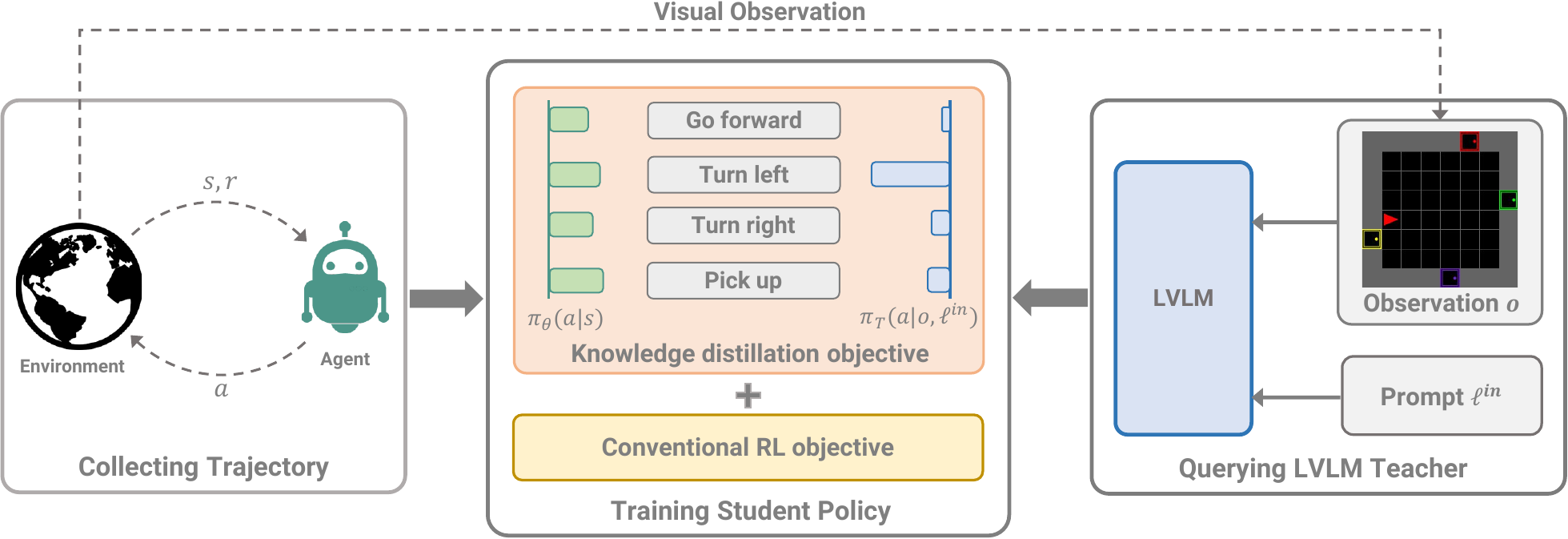}
    \caption{Given trajectories sampled by the student agent, we query the LVLM teacher during the RL update. The LVLM processes the visual observation and a textual prompt to generate the probability over a set of actions. The agent is then updated by optimizing both the conventional RL objective (\eg, PPO \cite{Schulman2017ProximalPO} or A2C \cite{Mnih2016AsynchronousMF}) and a knowledge distillation objective. Our proposed framework leverages the strengths of LVLMs, eliminating the need for handcrafted textual descriptors of states and significantly accelerating the student RL agent’s learning process.}
    \label{fig:overall}
    \vskip -0.15in
\end{figure*}

The proprosed knowledge distillation framework from the LVLM offers several key advantages. First, it eliminates the need for textual descriptors of states from the environment, as required in previous LLM-based approaches \cite{du2023guiding,singh2023progprompt,huang2022inner,hu2023enabling,carta2023grounding}, which often involve labor-intensive hand-engineering and may not always be readily available. Second, the student agent benefits from the LVLM's nuanced understanding of complex scenarios, enabling more strategic and informed decision-making while avoiding less meaningful exploration. This leads to significantly greater sample efficiency compared to learning from scratch. Finally, the distilled agent retains a high level of performance fidelity relative to its LVLM teacher while operating at a fraction of the computational cost, making it highly suitable for deployment in resource-constrained systems. To summarize, our main contributions are as follows: (1) We introduce LVLM2P, a knowledge distillation approach that leverages LVLM to reduce the sample complexity of RL-based agents in decision-making tasks. (2) Through extensive experiments on challenging grid-based tasks, we demonstrate that LVLM2P significantly reduces the sample complexity of $2.x$ times compared to the baselines. (3) We further show that the proposed framework can be seamlessly integrated into other RL algorithms. Additionally, we conduct ablation studies to understand the workings of LVLM2P.
\vspace{-0.7mm} 
\section{Related Works}
\noindent
\textbf{Utilizing Multimodal Foundation Models for RL.}
Many prior works, building on recent advancements in the advanced multimodal capabilities of pretrained LVLMs \cite{openai2023gpt4v,deepmind2023gemini, yang2024qwen2, TheC3}, have explored various methods for decision-making tasks. One line of research focuses on prompting techniques to assist in training RL agents, such as suggesting plans \cite{ahn2022can,singh2023progprompt,dalal2024plan}, generating reward signals \cite{kwon2023reward,rocamonde2023vision,wang2024rl}, or discovering meaningful skills \cite{wang2023voyager,du2023guiding,zhang2023bootstrap}. In contrast to these approaches, our method leverages the rich internalized knowledge of the LVLM through prompting techniques to accelerate the learning process of RL agents, thereby reducing the sample complexity of RL algorithms. Another line of work involves training additional heads to connect a frozen LVLM to the action space \cite{szot2023large,mu2024embodiedgpt,chen2024vision} or fine-tuning the entire LVLM \cite{zhai2024fine} to use it directly as a policy or planner. Unlike these methods, our approach avoids using large models as policies, making it more scalable and practical for resource-constrained systems.

\noindent
\textbf{Knowledge Distillation in RL. } Previous research has explored various techniques for transferring knowledge from existing controllers to smaller agents. For instance, \cite{Rusu2015PolicyD,parisotto2015actor} focus on distilling multiple pretrained policies into a unified network to create a single model capable of performing multiple tasks. On the other hand, several approaches leverage model-based controllers \cite{levine2013guided,nagabandi2018neural} or existing teacher agents \cite{da2020agents,agarwal2022reincarnating} to guide the exploration of RL agents during training, thereby expediting the learning process. Similarly, \cite{schmitt2018kickstarting} train RL agents by combining on-policy distillation with an RL objective, encouraging the student policy to imitate the teacher policy while simultaneously optimizing for long-term rewards. Our proposed method is inspired by this training paradigm \cite{schmitt2018kickstarting}; however, unlike previous work, we do not rely on pretrained teacher policies specific to the target task. Instead, we harness the internalized knowledge of a multimodal foundation model, which encompasses fundamental skills, to expedite the agent's learning process in solving a target task.

\section{Problem Formulation}
We formalize the decision-making problem as a Markov decision process (MDP), defined as $\mathcal{M} = (\mathcal{S}, \mathcal{A}, R, P, \gamma)$. The tuple consists of  states $s \in \mathcal{S}$, actions $a\in \mathcal{A}$, a reward function $r=R(s,a,s')$, transition probability $P(s'|s, a)$ and $\gamma \in [0, 1)$ is a discount factor. An RL agent takes actions based on a policy $\pi_{\theta}(a|s)$, parameterized by the parameter $\theta$. The objective of the RL agent is to maximize the expected returns $\mathbb{E}_\pi\left[\sum_{t=0}^{\infty}\gamma^tR(s_t, a_t, s_{t+1})\right]$. We additionally define the teacher policy as $\pi_{T}(a|o, \ell^{in})$, where $\ell^{in}$ is the linguistic prompt used to query the LVLM for instructional actions, and $o$ is the visual observation. Note that the policy operates on the environment's state rather than on the visual observation. The teacher policy directly leverages this image-based visual observation to understand the current state of the environment and produces stochastic actions, thus bypassing the need for hand-crafted textual descriptors of states \cite{du2023guiding,carta2023grounding}.

\section{Methodology}

\subsection{LVLM2P Framework}

We first introduce the overall framework of LVLM2P, as depicted in Fig. \ref{fig:overall}, which trains a lightweight student RL agent. This agent not only optimizes the usual RL objective (\ie, expected return) but also leverages a pre-trained LVLM teacher to accelerate training and enhance performance. 
The core idea of LVLM2P is to utilize a pre-trained LVLM as the teacher agent to suggest appropriate actions for visual observations within the trajectories \textit{collected} by the student. During policy updates, we then apply a knowledge distillation objective to align the student policy with the teacher policy on these trajectories. Our framework enables the student agent to freely interact with the environment based on its own ``knowledge'' and receive feedback from these interactions, correcting any potential errors stemming from the LVLM teacher agent. With the useful suggested actions from the LVLM teacher, the student policy avoids less meaningful exploration early on, enabling it to quickly focus on high-reward regions in the action and state space, thereby improve sample efficiency.

\subsection{LVLM as Teacher Agent}

\begin{figure}[H]
    \centering
    \begin{subfigure}{\columnwidth}
    \includegraphics[width=1\linewidth, trim=0cm 2.6cm 0cm 0cm, clip]{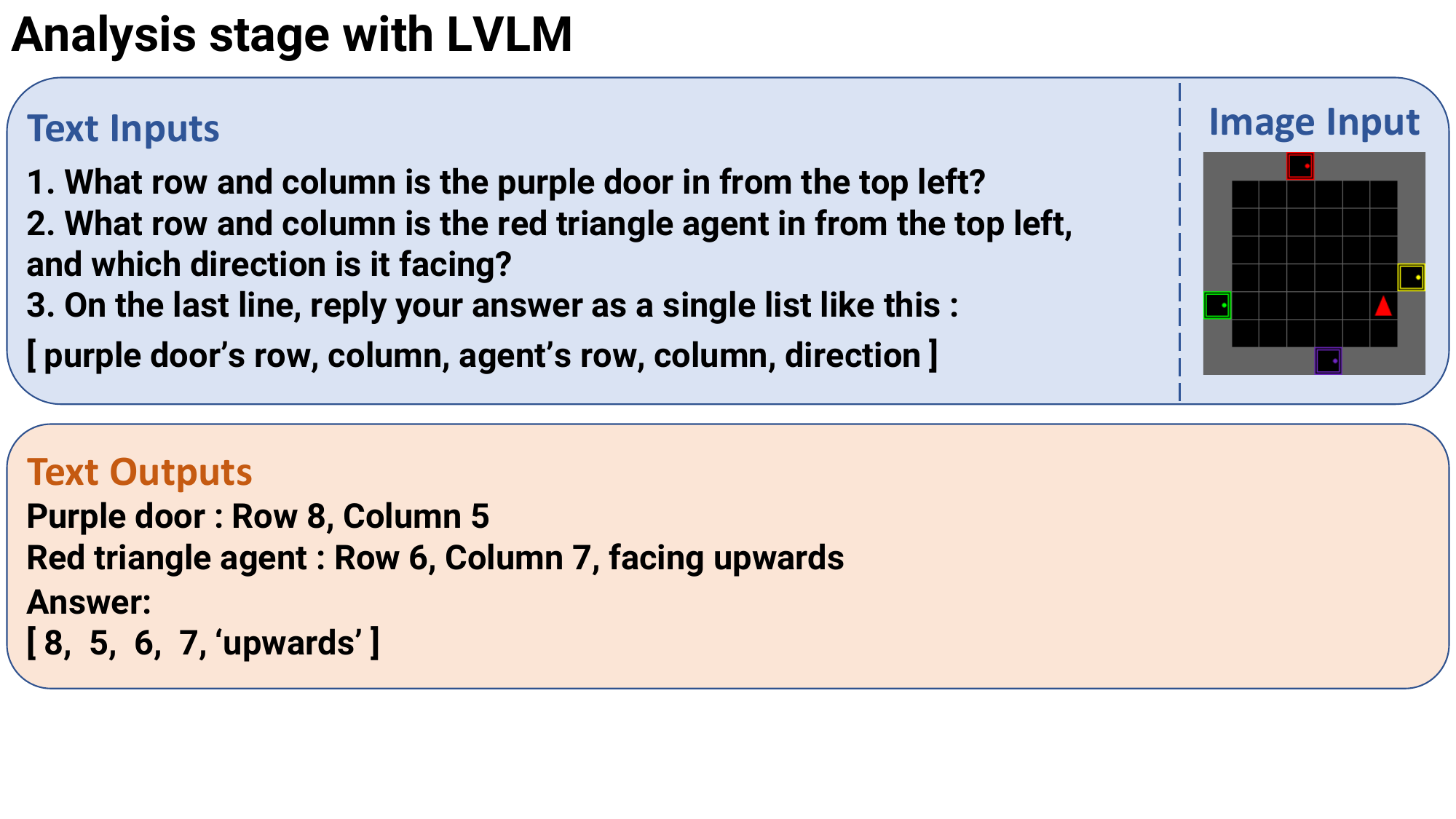}
    \end{subfigure}
    
    \begin{subfigure}{\columnwidth}   \includegraphics[width=\linewidth, trim=0cm 3.7cm 0cm 0cm, clip]{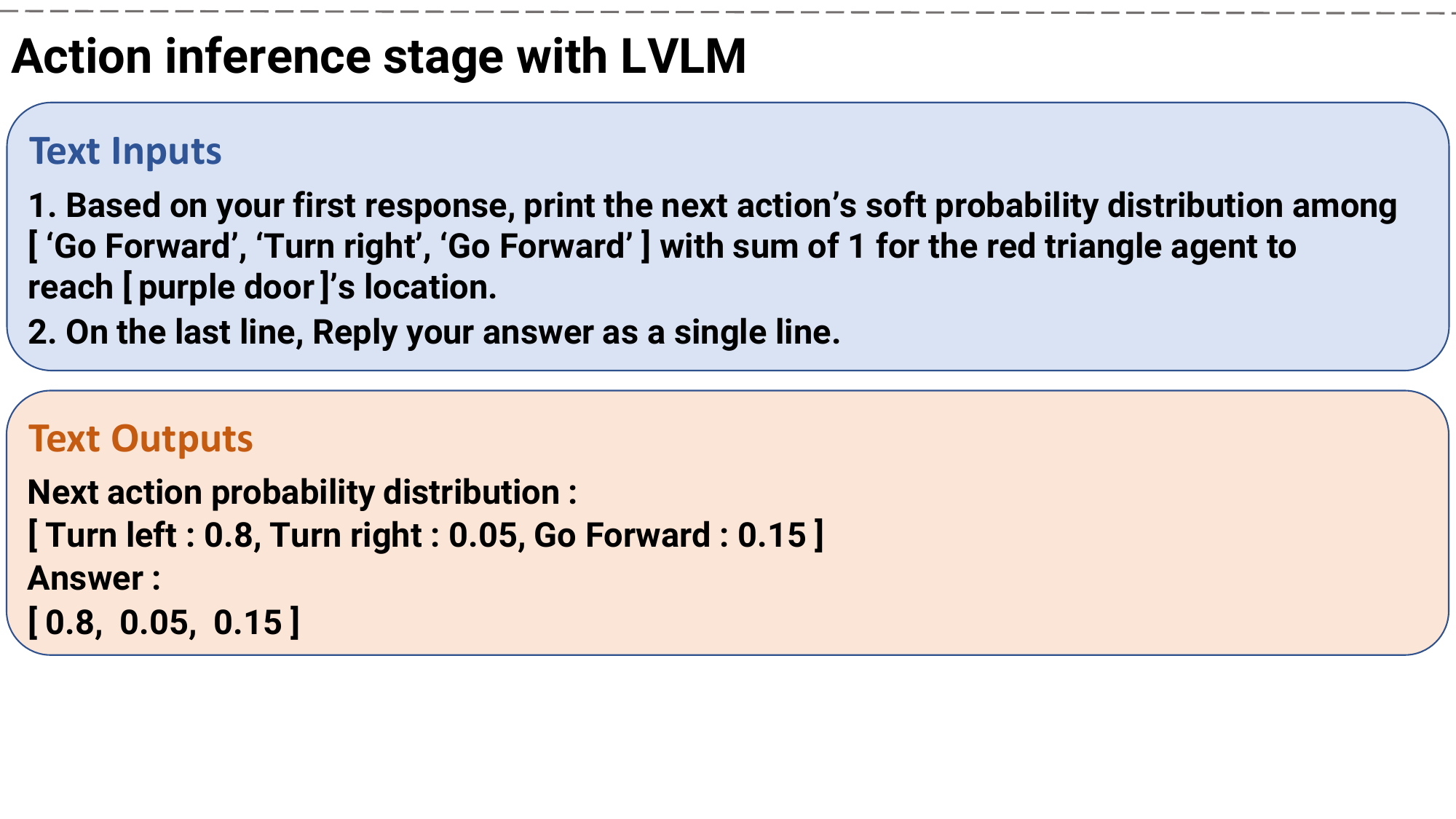}
    \end{subfigure}
    \caption{An example of a prompt in GoToDoor: In the first stage, the LVLM outputs key information such as the location and orientation of the agent (red triangle) and the target object (purple door). In the second stage, the LVLM is required to provide a numerical probability over possible actions.}
    \label{fig:prompt_example}
\end{figure}

Building on prior work using LLMs \cite{brown2020language,ahn2022can,carta2023grounding}, we provide carefully designed prompts, along with an image of the environment, to guide the behavior of the pre-trained LVLM in generating instructional actions. The querying process, illustrated in Fig. \ref{fig:prompt_example}, consists of two stages: an initial \textbf{analysis} stage followed by an \textbf{action inference} stage. In the \textbf{analysis} stage, given a visual observation from a trajectory, the LVLM is prompted to summarize key information in the scene, such as the coordinates and orientation of the agent and the target object. Then, in the \textbf{action inference} stage, we prompt the LVLM with the analysis from first stage to provide a numerical probability over a set of possible actions. In this second stage, we also include few-shot examples to ensure consistency in the LVLM's responses. Additionally, we use soft probabilities over actions instead of a single action (\ie, one-hot encoding), which has been shown to improve sample efficiency and prevent the student model from overfitting during knowledge distillation \cite{hinton2015distilling, muller2019does}. This process is then repeated for all observations within a trajectory collected by the student agent.

\subsection{Training Process of Student Policy}

In this paper, we apply LVLM2P to on-policy RL algorithms such as PPO \cite{Schulman2017ProximalPO} and A2C \cite{Mnih2016AsynchronousMF}, though our approach can easily be extended to other RL algorithms as well. During training, the student policy $\pi_{\theta}$ is trained to optimize the following objective:
\begin{equation}\label{eq:training_loss}
\mathcal{L}(\pi_{\theta}) = \mathcal{L}_{RL}(\pi_{\theta}) + \lambda D_{KL}(\pi_T(\cdot | o, \ell^{in}) \| \pi_{\theta}(\cdot | s))
\end{equation}
where $\mathcal{L}_{RL}(\pi_{\theta})$ represents the conventional RL objective, which encourages the student policy to maximize long-term rewards, $D_{KL}$ denotes the KL divergence used for knowledge distillation, and $\lambda$ is the coefficient that balances the two objectives. The procedure for training LVLM2P with an RL algorithm is summarized in Algorithm \ref{alg:lvlm2p}. 

\vspace{-5.5mm} 
\begin{algorithm}[H]
    \small
    \caption{LVLM2P training algorithm}
    \label{alg:lvlm2p}
    \begin{algorithmic}[1]
    \STATE \textbf{Input}: A pre-trained LVLM, the coefficient $\lambda$, $N$ training steps, $K$ policy gradient update steps, trajectory length $T$
    \STATE \textbf{Initialize}: Student policy $\pi_{\theta}$
    \FOR{$i = 1$ to $N$}
        \STATE // \texttt{Collecting trajectories}
        \STATE $\mathcal{B} \leftarrow \emptyset$
        \FOR{$t = 1$ to $T$}
        \STATE Obtain $s_{t+1}$, $o_{t+1}$, $r_t$ by taking $a_t \sim \pi_{\theta} (\cdot|s_t)$
      \STATE Update buffer $\mathcal{B} \leftarrow \mathcal{B} \cup \{(s_t, a_t, s_{t+1}, r_t, o_t)\}$
    \ENDFOR
    \STATE // \texttt{Student policy update}
        \FOR{$k = 1$ to $K$}
            \STATE Sample random batch $\{(s_t, a_t, s_{t+1}, r_t, o_t)_j\}_{j=1}^B \sim \mathcal{B}$
            \STATE Query LVLM teacher for probability $p_t$ over actions: $p_t \leftarrow \pi_T(\cdot|o_t, \ell^{in})$ for every $o_t$ inthe  sampled batch
            \STATE Update student policy:
            \begin{center}
                $\theta \leftarrow \theta + \alpha\nabla_{\theta}\left(\mathcal{L}_{RL}(\pi_{\theta}) + \lambda D_{KL}(p_t \| \pi_{\theta}(\cdot|s_t))\right)$
            \end{center}
        \ENDFOR
    \ENDFOR 
    \end{algorithmic}
\end{algorithm} 
\vspace{-3.5mm} 
\noindent

\begin{figure*}[t]
    \centering
    \begin{subfigure}{0.85\textwidth}
    \hspace{-0.315cm}
\includegraphics[width=\linewidth, trim=0cm 0cm 0cm 0.24cm, clip]{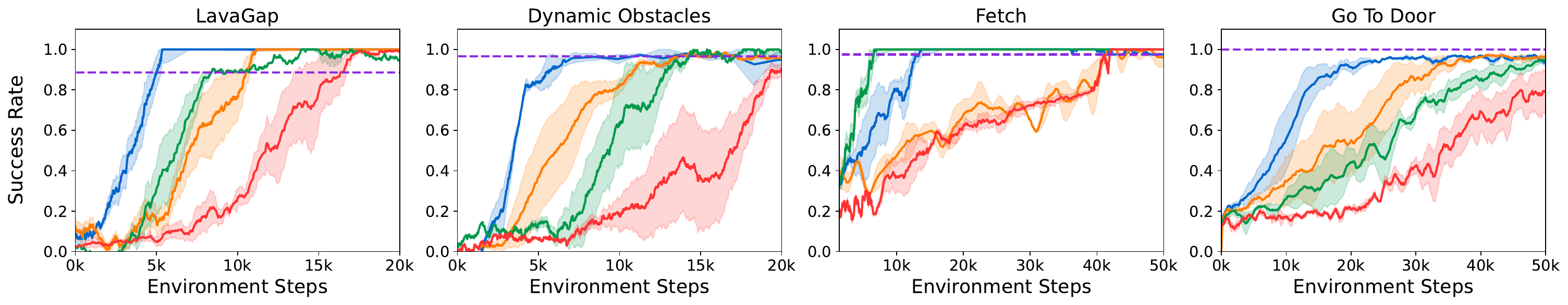}
    \end{subfigure}
    
    \begin{subfigure}{0.85\textwidth}   
    \includegraphics[width=\linewidth, trim=0cm 0cm 0cm 0cm, clip]{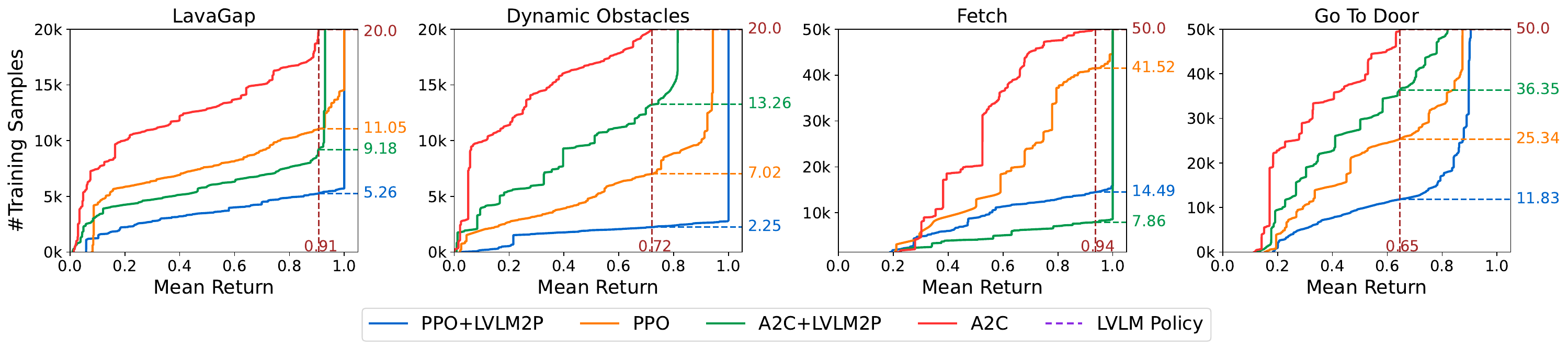}
    
    \end{subfigure}
    \caption{\textit{Top}: The mean success rate, along with the standard deviation range, across all four environments (the higher the better). \textit{Bottom}: Number of training samples needed with respect to mean return for all four environment (the lower the better). Each method was evaluated using three random seeds.}
    \label{fig:results}
    \vskip -0.17in
\end{figure*}
\vspace{-1mm}
\section{Experiments}

\noindent
\textbf{Experimental Setup.} To investigate whether our LVLM2P method enhances the sample efficiency of the base RL algorithm in decision-making tasks, we have adopted four sparse reward tasks from the MiniGrid\cite{chevalier2024minigrid} environments: 
\textbf{LavaGap}: Agents navigate through a narrow gap surrounded by deadly lava to emphasize safe exploration. 
\textbf{Dynamic Obstacles}: Challenges agents to avoid moving obstacles in an empty room, testing dynamic obstacle avoidance under partial observability. 
\textbf{Fetch}: Agents must correctly identify and retrieve specific objects based on textual instructions, focusing on precise object recognition and manipulation. 
\textbf{GoToDoor}: Requires agents to locate and approach the correct door as indicated by a textual cue, assessing their navigation skills in response to the mission instructions.
We utilized Gemini-1.5-Flash with few-shot prompts as the teacher policy, as it demonstrated the best trade-off between inference time and task accuracy in the extensive decision-making experiments presented in \cite{reid2024gemini}. The LVLM teacher policy receives a fully-observable view of the environment along with a prompt input $\ell^{in}$. For the student policy, we establish baselines using PPO\cite{Schulman2017ProximalPO} and A2C\cite{Mnih2016AsynchronousMF} as the underlying RL algorithms for training student policies.

\noindent
\textbf{Experimental Results.} As depicted in first row of the Fig. \ref{fig:results}, the LVLM policy achieves an average success rate of about 0.96, indicative of its outstanding decision-making performance and  its effectiveness in transferring knowledge to the student models. Our experiments further reveal that guiding the student policy through knowledge distillation from the LVLM teacher policy significantly enhances decision-making capabilities. In all four tasks, both A2C and PPO intergrated with LVLM2P converge more rapidly than their vanilla counterparts, demonstrating that the incorporation of knowledge from the LVLM leads to improved learning efficacy. In second row, our analysis in the LavaGap environment shows that A2C and PPO, when integrated with LVLM2P, required significantly fewer training samples to achieve a mean return of 0.91—9.18k and 5.26k samples respectively, making them 2.18 and 2.11 times more sample-efficient than their vanilla counterparts. Across all environments, the enhancements with LVLM2P improved sample efficiency by an average of 2.56 times for PPO and 2.86 times for A2C.

\begin{figure}[t]
    \centering
    \begin{subfigure}{\columnwidth}
    \includegraphics[width=1\linewidth, trim=0cm 0cm 0cm 0.4cm, clip]{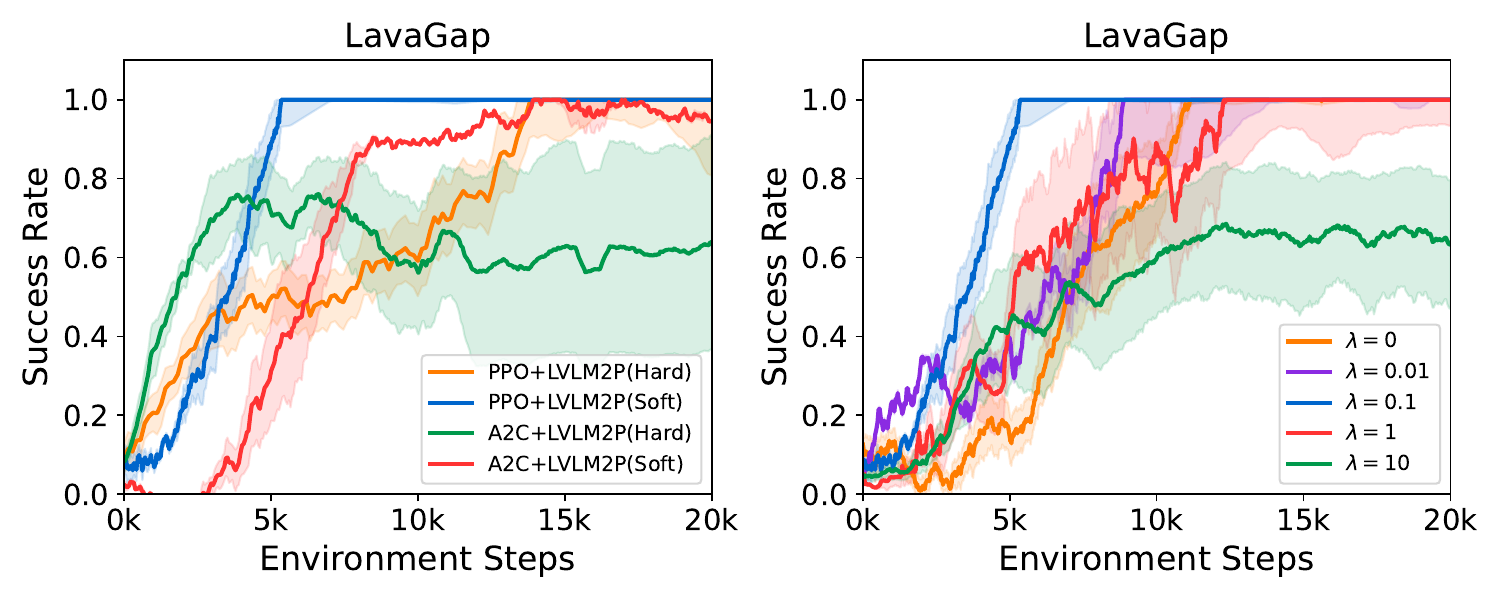}
    \end{subfigure}

    \caption{Ablation study of LVLM2P : Comparison of hard and soft probabilities from the LVLM teacher policy (left) and the impact of hyper-parameter \(\lambda\) on success rate results (right).
Each method was evaluated using three random seeds.}
    \label{fig:ablation study}
    \vskip -0.20in
\end{figure}

\textbf{Ablation Study}
We conduct an ablation study to analyze the components of our LVLM2P method in the LavaGap environment. Fig. \ref{fig:ablation study} demonstrates that LVLM2P with soft probability distributions from the LVLM teacher policy outperforms hard distributions, as demonstrated in \cite{hinton2015distilling}. This indicates that soft probability distribution provide richer information for learning by encompassing probabilities across all action classes, thereby providing a more nuanced understanding. This contrasts with hard distribution, which focus solely on the action class output from the teacher model and can result in extreme error signals on incorrect predictions.
Additionally, we investigates the impact of the $\lambda$ coefficient on PPO with LVLM2P. Results show optimal success rates with $\lambda = 0.01$, surpassing those lacking no teacher policy influence ($\lambda = 0$), and success rates decline as $\lambda$ increases beyond 1.

%

\vspace{1mm}
\section{Conclusion} 
\vspace{1mm}
In this paper, we introduced LVLM2P, a novel framework that integrates LVLMs with RL for decision-making tasks. By leveraging the reasoning capabilities of LVLM, our approach enables the development of highly efficient RL agents. significantly enhances sample efficiency, as shown in extensive experiments. Additionally, by using LVLM as a teacher capable of inferring visual information, our framework eliminates the need for handcrafted textual descriptions of the environment, simplifying deployment. Importantly, LVLM2P is versatile, seamlessly integrating with various RL algorithms, proving its adaptability and effectiveness across a wide range of tasks. We believe our work lays the foundation for extending LVLM-based frameworks to future continuous control tasks.


\newpage   

\bibliographystyle{IEEEtran}
\bibliography{references}

\end{document}